\documentclass[final]{cvpr}

\usepackage{times}
\usepackage{epsfig}
\usepackage{graphicx}
\usepackage{amsmath}
\usepackage{amssymb}

\usepackage{booktabs}
\usepackage{subcaption}
\usepackage{multirow}

\usepackage{color}
\usepackage{xcolor}
\definecolor{citecolor}{HTML}{0071bc}
\usepackage[pagebackref=true,breaklinks=true,colorlinks,citecolor=citecolor,bookmarks=false]{hyperref}

\begin{document}

%\pagestyle{empty}

%%%%%%%%% TITLE
\title{Learning Continuous Image Representation with Local Implicit Image Function}

\author{Yinbo Chen\\
UC San Diego
% For a paper whose authors are all at the same institution,
% omit the following lines up until the closing ``}''.
% Additional authors and addresses can be added with ``\and'',
% just like the second author.
% To save space, use either the email address or home page, not both
\and
Sifei Liu\\
NVIDIA
\and
Xiaolong Wang\\
UC San Diego
}

\maketitle
%\thispagestyle{empty}

%%%%%%%%% ABSTRACT
\begin{abstract}
How to represent an image? While the visual world is presented in a continuous manner, machines store and see the images in a discrete way with 2D arrays of pixels. In this paper, we seek to learn a continuous representation for images. Inspired by the recent progress in 3D reconstruction with implicit neural representation, we propose Local Implicit Image Function (LIIF), which takes an image coordinate and the 2D deep features around the coordinate as inputs, predicts the RGB value at a given coordinate as an output. Since the coordinates are continuous, LIIF can be presented in arbitrary resolution. To generate the continuous representation for images, we train an encoder with LIIF representation via a self-supervised task with super-resolution. The learned continuous representation can be presented in arbitrary resolution even extrapolate to $\times 30$ higher resolution, where the training tasks are not provided. We further show that LIIF representation builds a bridge between discrete and continuous representation in 2D, it naturally supports the learning tasks with size-varied image ground-truths and significantly outperforms the method with resizing the ground-truths. Our project page with code is at \url{https://yinboc.github.io/liif/}.
\end{abstract}

%%%%%%%%% BODY TEXT

\section{Introduction}

Our visual world is continuous. However, when a machine tries to process a scene, it will usually need to first store and represent the images as 2D arrays of pixels, where the trade-off between complexity and precision is controlled by resolution. While the pixel-based representation has been successfully applied in various computer vision tasks, they are also constrained by the resolution. For example, a dataset is often presented by images with different resolutions. If we want to train a convolutional neural network, we will usually need to resize the images to the same size, which may sacrifice fidelity. Instead of representing an image with a fixed resolution, we propose to study a continuous representation for images. By modeling an image as a function defined in a continuous domain, we can restore and generate the image in arbitrary resolution if needed. 

How do we represent an image as a continuous function? Our work is inspired by the recent progress in implicit neural representation~\cite{park2019deepsdf,mescheder2019occupancy,chen2019learning,saito2019pifu,jiang2020local,sitzmann2020implicit} for 3D shape reconstruction. The key idea of implicit neural representation is to represent an object as a function that maps coordinates to the corresponding signal (e.g. signed distance to a 3D object surface, RGB value in an image), where the function is parameterized by a deep neural network. To share knowledge across instances instead of fitting individual functions for each object, encoder-based methods~\cite{mescheder2019occupancy,chen2019learning,sitzmann2020implicit} are proposed to predict latent codes for different objects, then a decoding function is shared by all the objects while it takes the latent code as an additional input to the coordinates. Despite its success in 3D tasks~\cite{saito2019pifu,saito2020pifuhd}, previous encoder-based methods of implicit neural representation only succeeded in representing simple images such as digits~\cite{chen2019learning}, but failed to represent natural images with high fidelity~\cite{sitzmann2020implicit}.

\begin{figure}
    \centering
    \includegraphics[width=\linewidth]{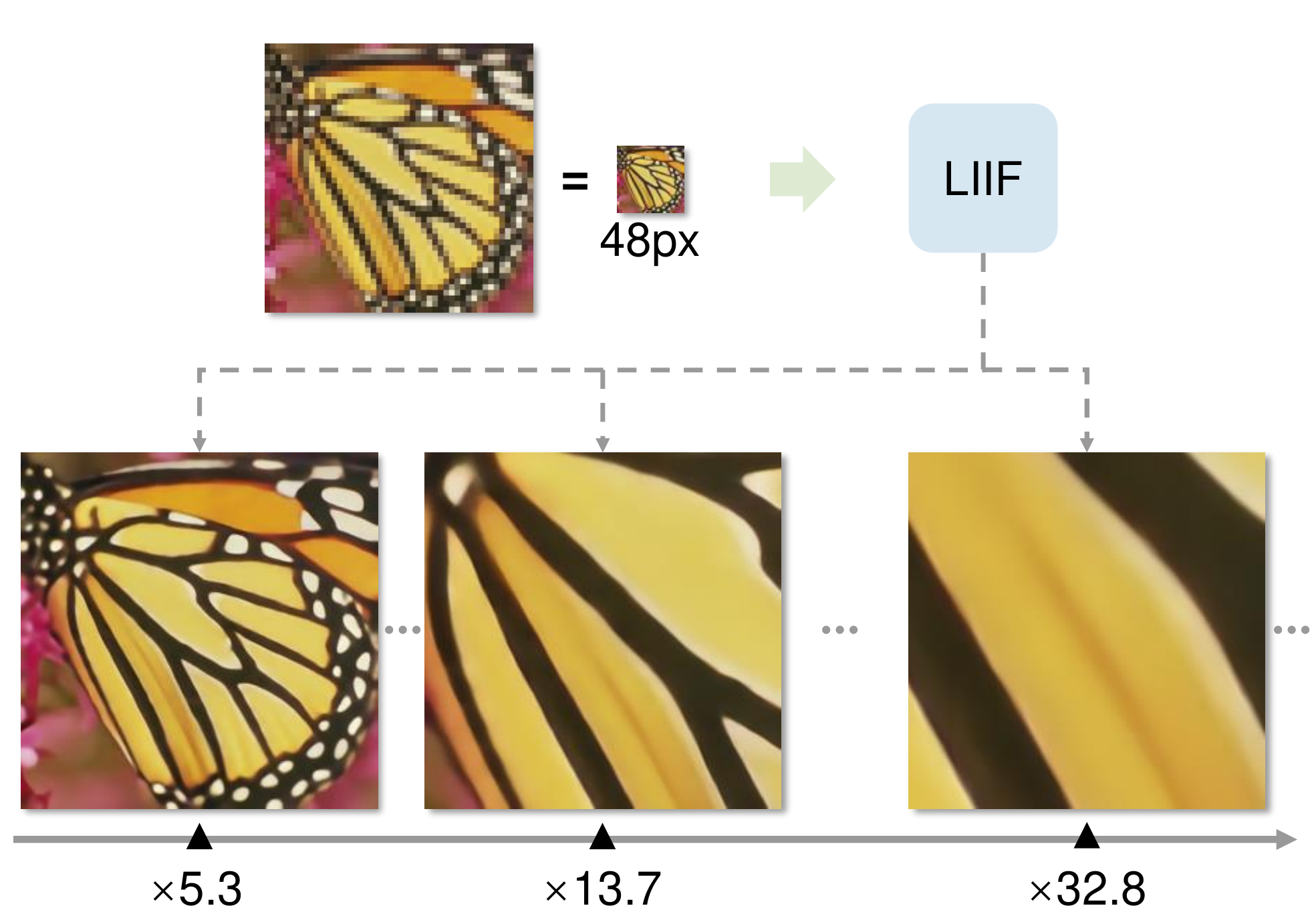}
    \caption{Local Implicit Image Function (LIIF) represents an image in continuous domain, which can be presented in arbitrary high resolution.}
    %\vspace{-1em}
    \label{fig:teaser}
\end{figure}

In this paper, we propose the Local Implicit Image Function (LIIF) for representing natural and complex images in a continuous manner. In LIIF, an image is represented as a set of latent codes distributed in spatial dimensions. Given a coordinate, the decoding function takes the coordinate information and queries the local latent codes around the coordinate as inputs, then predicts the RGB value at the given coordinate as an output. Since the coordinates are continuous, LIIF can be presented in arbitrary resolution.

To generate such continuous representation for pixel-based images, since we hope the generated continuous representation can generalize to higher precision than the input image, we train an encoder with the LIIF representation via a self-supervised task with super-resolution, where the input and ground-truth are provided in continuously changing up-sampling scales. In this task, take a pixel-based image as an input, the encoded LIIF representation is trained to predict a higher resolution counterpart of the input. While most of the previous works on image super-resolution~\cite{dong2015image,ledig2017photo,lim2017enhanced,lai2017deep} focus on learning an up-sampling function for specific scales in a convolution-deconvolution framework, LIIF representation is continuous, and we show it can be presented in arbitrary high resolution, that can even extrapolate to $\times 30$ higher resolution where the training tasks are not provided.

We further demonstrate that LIIF builds a bridge between discrete and continuous representation in 2D. In the learning tasks with size-varied image ground-truths, LIIF can naturally exploit the information provided in different resolutions. Previous methods with fixed-size output usually need to resize all the ground-truths to the same size for training, which may sacrifice fidelity. Since the LIIF representation can be presented in arbitrary resolution, it can be trained in an end-to-end manner without resizing ground-truths, which achieves significantly better results than the method with resizing the ground-truths.

Our contributions include: (i) A novel method for representing natural and complex images continuously; (ii) LIIF representation allows extrapolation to even $\times 30$ higher resolution which is not presented during training time; (iii) We show LIIF representation is effective for the learning tasks with size-varied image ground-truths.

\section{Related Work}

\textbf{Implicit neural representation.} In implicit neural representation, an object is usually represented as a multilayer perceptron (MLP) that maps coordinates to signal. This idea has been widely applied in modeling 3D object shapes~\cite{chen2019learning,michalkiewicz2019implicit,atzmon2020sal,gropp2020implicit}, 3D surfaces of the scene~\cite{sitzmann2019scene,jiang2020local,peng2020convolutional,chabra2020deep} as well as the appearance of the 3D structure~\cite{oechsle2019texture,niemeyer2020differentiable,mildenhall2020nerf}. For example, Mildenhall et al.~\cite{mildenhall2020nerf} propose to perform novel view synthesis by learning an implicit representation for a specific scene using multiple image views. Comparing to explicit 3D representations such as voxel, point cloud, and mesh, the continuous implicit representation can capture the very fine details of the shape with a small number of parameters. Its differentiable property also allows back-propagation through the model for neural rendering~\cite{sitzmann2019scene} . 

\textbf{Learning implicit function space.} Instead of learning an independent implicit neural representation for each object, recent works share a function space for the implicit representations of different objects. Typically, a latent space is defined where each object corresponds to a latent code. The latent code can be obtained by optimization with an auto-decoder~\cite{park2019deepsdf,chen2019learning}. For example, Park et al.~\cite{park2019deepsdf} propose to learn a Signed Distance Function (SDF) for each object shape and different SDFs can be inferred by changing the input latent codes. Recent work from Sitzmann et al.~\cite{sitzmann2020metasdf} also proposes a meta-learning-based method for sharing the function space. Instead of using auto-decoder, our work adopts the auto-encoder architecture~\cite{mescheder2019occupancy,chen2019learning,saito2019pifu,saito2020pifuhd,xu2019disn}, which gives an efficient and effective manner for sharing knowledge between a large variety of samples. For example, Mescheder et al.~\cite{mescheder2019occupancy} propose to estimate a latent code given an image as input, and use an occupancy function conditioning on this latent code to perform 3D reconstruction for the input object.

Despite the success of implicit neural representation in 3D tasks, its applications in representing images are relatively underexplored. Early works~\cite{stanley2007compositional,mordvintsev2018differentiable} parameterize 2D images with compositional pattern producing networks. Chen et al.~\cite{chen2019learning} explore 2D shape generation from latent space for simple digits. Recently, Sitzmann et al.~\cite{sitzmann2020implicit} observe that previous implicit neural representation parameterized by MLP with ReLU~\cite{nair2010rectified} is incapable of representing fine details of natural images. They replace ReLU with periodic activation
functions (sinusoidal) and demonstrates it can model the natural images in higher quality. However, none of these approaches can represent natural and complex images with high fidelity when sharing the implicit function space, while it is of limited generalization to higher precision if not to share the implicit function space. Related to recent works~\cite{saito2019pifu,genova2020local,chibane2020implicit,peng2020eccv,jiang2020local} on 3D implicit neural representation, LIIF representation is based on local latent codes, which can recover the fine details of natural and complex images. Similar formulations have been recently proposed for 3D reconstruction~\cite{jiang2020local} and super-resolving physics-constrained solution~\cite{jiang2020meshfreeflownet}. Different from these works, LIIF focuses on learning continuous image representation and has image-specific design choices (e.g. cell decoding).

\textbf{Image generation and super-resolution.} Our work is related to the general image-to-image translation tasks~\cite{zhang2016colorful,pix2pix2017,zhu2017unpaired,hoffman2018cycada,denton2015deep}, where one image is given as input and it is translated to a different domain or format. For example, Isola et al.~\cite{pix2pix2017} propose conditional GANs~\cite{goodfellow2014generative} to perform multiple image translation tasks. Unlike the deconvolution-based approaches, LIIF representation supports performing realistic and high-resolution image generation by independently querying the pixels at different coordinates from the generated implicit representation. While LIIF is useful for general purposes, in this paper, we perform experiments on generating high-resolution images given low-resolution inputs, which is related to the image super-resolution tasks~\cite{chang2004super,timofte2013anchored,tai2017memnet,dong2014learning,ledig2017photo,lim2017enhanced,lai2017deep,zhang2018residual,zhang2020deep}. For example, Lai et al.~\cite{lai2017deep} propose a Laplacian Pyramid Network to progressively reconstruct the image. While related, we stress that this work on learning continuous representation is different from the traditional super-resolution setting. Most previous super-resolution models are designed for up-sampling with a specific scale (or a fixed set of scales), while our goal is to learn a continuous representation that can be presented in arbitrary high resolution. In this respect, our work is more related to MetaSR~\cite{hu2019meta} on magnification-arbitrary super-resolution. Their method generates a convolutional up-sampling layer by its meta-network, while it can perform arbitrary up-sampling in its training scales, it has limited performance on generalizing to the larger-scale synthesis that is out of training distribution. LIIF representation, on the other hand, when trained with tasks from $\times 1$ to $\times 4$, can generate $\times 30$ higher resolution image based on the continuous representation in one forward pass.

\section{Local Implicit Image Function}

\begin{figure}
    \centering
    \includegraphics[width=.62\linewidth]{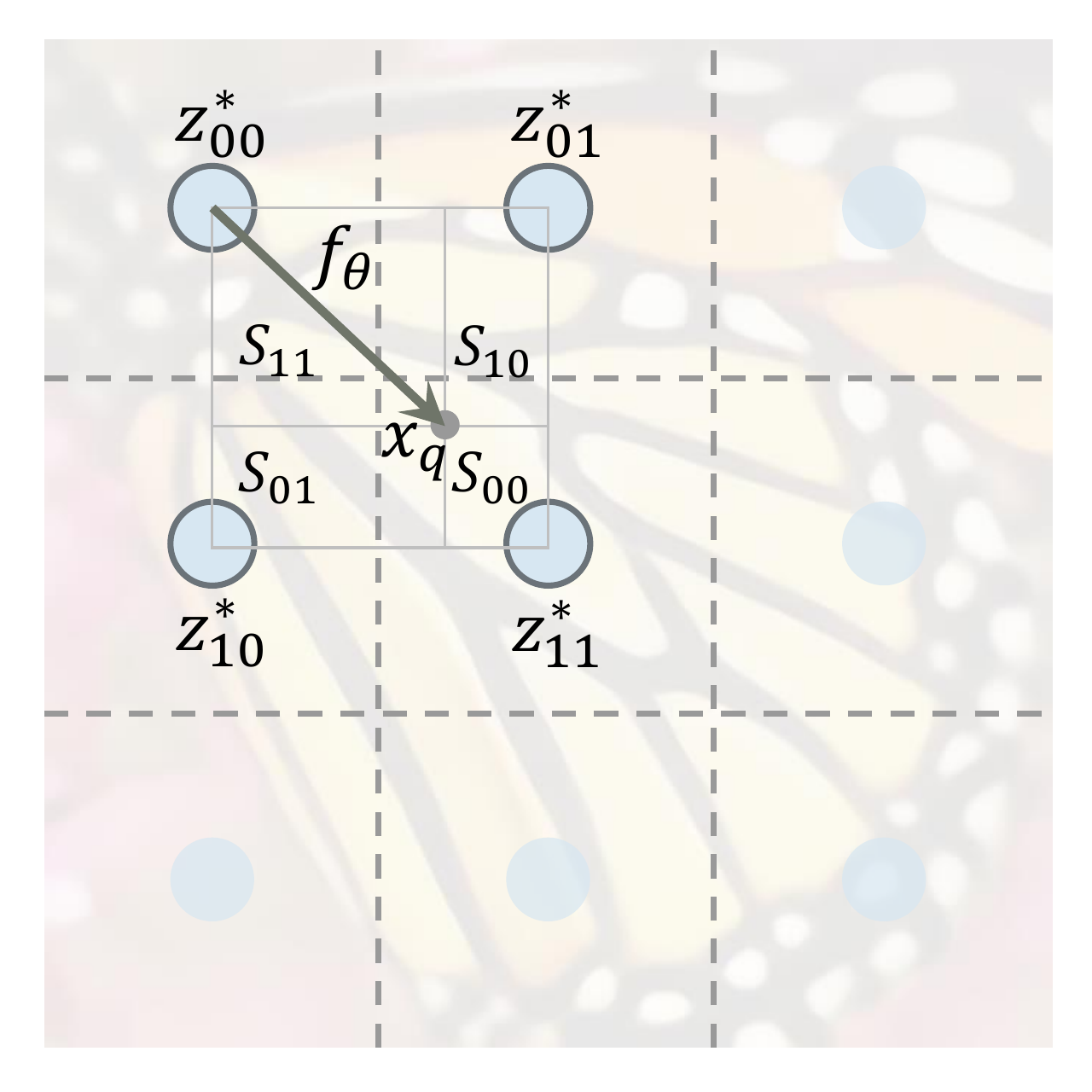}
    \caption{\textbf{LIIF representation with local ensemble.} A continuous image is represented as a 2D feature map with a decoding function $f_{\theta}$ shared by all the images. The signal is predicted by ensemble of the local predictions, which guarantees smooth transition between different areas.}
    \label{fig:liif}
\end{figure}

In LIIF representation, each continuous image $I^{(i)}$ is represented as a 2D feature map $M^{(i)} \in \mathbb{R}^{H\times W\times D}$. A decoding function $f_\theta$ (with $\theta$ as its parameters) is shared by all the images, it is parameterized as a MLP and takes the form:
\begin{equation}
    s = f_\theta(z, x),
    \label{eq:liif-f}
\end{equation}
where $z$ is a vector, $x\in \mathcal{X}$ is a 2D coordinate in the  continuous image domain, $s\in \mathcal{S}$ is the predicted signal (i.e. the RGB value). In practice, we assume the range of $x$ is $[0, 2H]$ and $[0, 2W]$ for two dimensions. With a defined $f_\theta$, each vector $z$ can be considered as representing a function $f_\theta(z,\cdot): \mathcal{X} \mapsto \mathcal{S}$, i.e. a function that maps coordinates to RGB values. We assume the $H\times W$ feature vectors (we call them latent codes from now on) of $M^{(i)}$ are evenly distributed in the 2D space of the continuous image domain of $I^{(i)}$ (e.g. blue circles in Figure~\ref{fig:liif}), then we assign a 2D coordinate to each of them. For the continuous image $I^{(i)}$, the RGB value at coordinate $x_q$ is defined as
\begin{equation}
    I^{(i)}({x_q}) = f_\theta(z^*, x_q - v^*),
    \label{eq:liif-local}
\end{equation}
where $z^*$ is the nearest (Euclidean distance) latent code from $x_q$ in $M^{(i)}$,  $v^*$ is the coordinate of latent code $z^*$ in the image domain. Take Figure~\ref{fig:liif} as an example, $z^*_{11}$ is the $z^*$ for $x_q$ in our current definition, while $v^*$ is defined as the coordinate for $z^*_{11}$.

As a summary, with a function $f_\theta$ shared by all the images, a continuous image is represented as a 2D feature map $M^{(i)} \in \mathbb{R}^{H\times W\times D}$ which is viewed as $H\times W$ latent codes evenly spread in the 2D domain. Each latent code $z$ in $M^{(i)}$ represents a local piece of the continuous image, it is responsible for predicting the signal of the set of coordinates that are closest to itself.

\vspace{-1em}
\paragraph{Feature unfolding.} To enrich the information contained in each latent code in $M^{(i)}$, we apply feature unfolding to $M^{(i)}$ and get $\hat{M}^{(i)}$. A latent code in $\hat{M}^{(i)}$ is the concatenation of the $3\times 3$ neighboring latent codes in $M^{(i)}$. Formally, the feature unfolding is defined as
\begin{equation}
    \hat{M}^{(i)}_{jk} = \textrm{Concat}(\{M^{(i)}_{j+l,k+m}\}_{l,m\in\{-1,0,1\}}),
\end{equation}
where Concat refers to concatenation of a set of vectors, $M^{(i)}$ is padded by zero-vectors outside its border. After feature unfolding, $\hat{M}^{(i)}$ replaces $M^{(i)}$ for any computation. For simplicity, we will only use the notation $M^{(i)}$ in the following sections regardless of feature unfolding. 

\vspace{-1em}
\paragraph{Local ensemble.} An issue in Eq~\ref{eq:liif-local} is the discontinuous prediction. Specifically, since the signal prediction at $x_q$ is done by querying the nearest latent code $z^*$ in $M^{(i)}$, when $x_q$ moves in the 2D domain, the selection of $z^*$ can suddenly switch from one to another (i.e. the selection of nearest latent code changes). For example, it happens when $x_q$ crossing the dashed lines in Figure~\ref{fig:liif}. Around those coordinates where the selection of $z^*$ switches, the signal of two infinitely close coordinates will be predicted from different latent codes. As long as the learned function $f_\theta$ is not perfect, discontinuous patterns can appear at these borders where $z^*$ selection switches.

To address this issue, as shown in Figure~\ref{fig:liif}, we extend Eq~\ref{eq:liif-local} to:
\begin{equation}
    I^{(i)}({x_q}) = \sum_{t\in\{00,01,10,11\}} \frac{S_t}{S} \cdot f_\theta(z^*_t, x_q - v^*_t),
\end{equation}
where $z^*_t (t\in\{00,01,10,11\})$ are the nearest latent code in top-left, top-right, bottom-left, bottom-right sub-spaces, $v^*_t$ is the coordinate of $z^*_t$, $S_t$ is the area of the rectangle between $x_q$ and $v^*_{t'}$ where $t'$ is diagonal to $t$ (i.e. 00 to 11, 10 to 01). The weights are normalized by $S=\sum_t S_t$. We consider the feature map $M^{(i)}$ to be mirror-padded outside the borders, so that the formula above also works for coordinates near the borders.

Intuitively, this is to let the local pieces represented by local latent codes overlap with its neighboring pieces so that at each coordinate there are four latent codes for independently predicting the signal. These four predictions are then merged by voting with normalized confidences, which are proportional to the area of the rectangle between the query point and its nearest latent code's diagonal counterpart, thus the confidence gets higher when the query coordinate is closer. It achieves continuous transition at coordinates where $z^*$ switches (e.g. dashed lines in Figure \ref{fig:liif}).

\begin{figure}
    \centering
    \includegraphics[width=.8\linewidth]{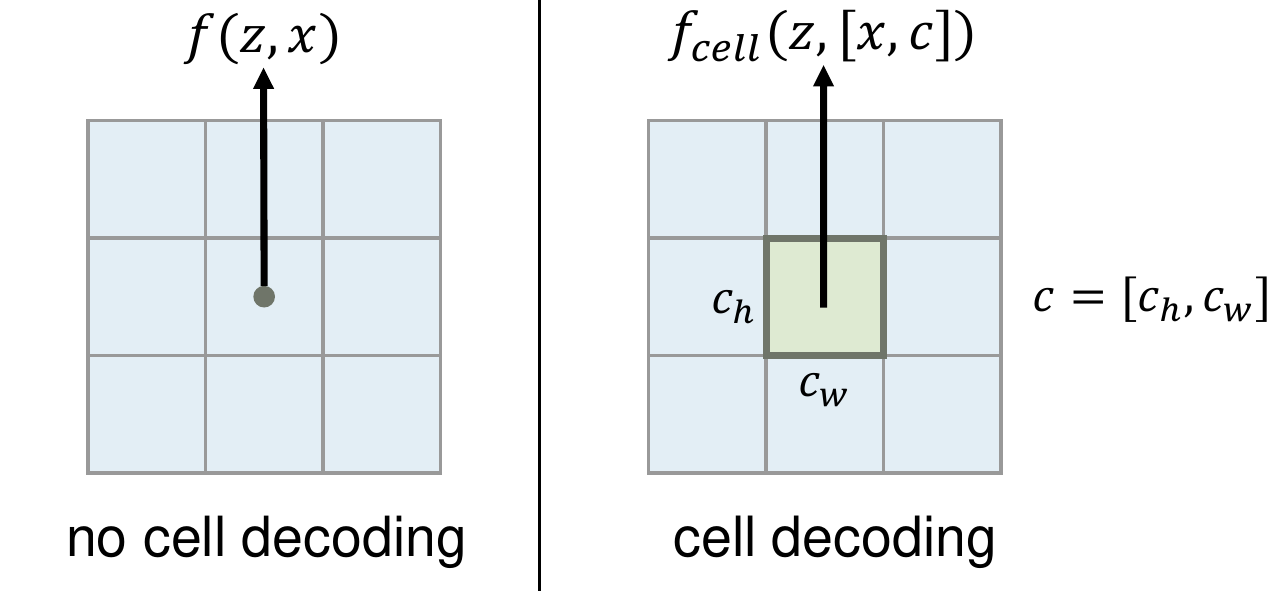}
    \caption{\textbf{Cell decoding.} With cell decoding, the decoding function takes the shape of the query pixel as an additional input and predicts the RGB value for the pixel.}
    \label{fig:cell}
\end{figure}

\vspace{-1em}
\paragraph{Cell decoding.} In practice, we want that the LIIF representation can be presented as the pixel-based form in arbitrary resolution. Suppose the desired resolution is given, a straight-forward way is to query the RGB values at the coordinates of pixel centers in the continuous representation $I^{(i)}(x)$. While this can already work well, it may not be optimal since the predicted RGB value of a query pixel is independent of its size, the information in its pixel area is all discarded except the center value.

To address this issue, we add cell decoding as shown in Figure~\ref{fig:cell}. We reformulate $f$ (omit $\theta$) in Eq~\ref{eq:liif-f} as $f_{cell}$ with the form
\begin{equation}
    s = f_{cell}(z, [x, c]),
\end{equation}
where $c=[c_h,c_w]$ contains two values that specify the height and width of the query pixel, $[x, c]$ refers to the concatenation of $x$ and $c$. The meaning of $f_{cell}(z, [x, c])$ can be interpreted as: what the RGB value should be, if we render a pixel centered at coordinate $x$ with shape $c$. As we will show in the experiments, having an extra input $c$ can be beneficial when presenting the continuous representation in a given resolution.

\begin{figure}
    \centering
    \includegraphics[width=\linewidth]{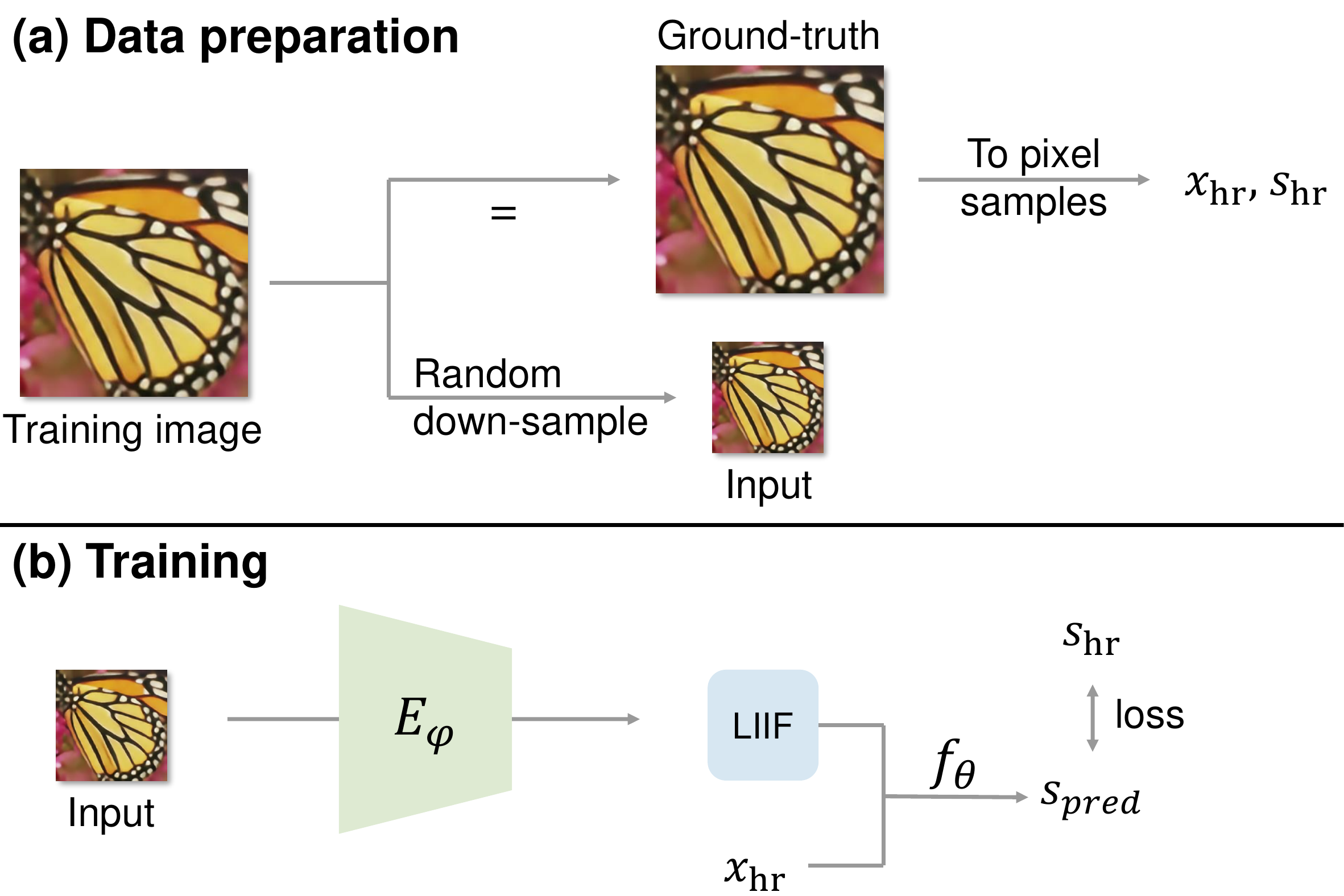}
    \caption{\textbf{Learning to generate continuous representation for pixel-based images.} An encoder is jointly trained with the LIIF representation in a self-supervised super-resolution task, in order to encourage the LIIF representation to maintain high fidelity in higher resolution.}
    \label{fig:learn-liif}
\end{figure}

\begin{figure*}[]
    \centering
    \begin{tabular}{ccccc}
        \includegraphics[width=4.4mm]{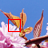} \includegraphics[width=17.6mm]{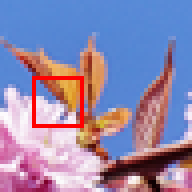} &
        \includegraphics[width=33mm]{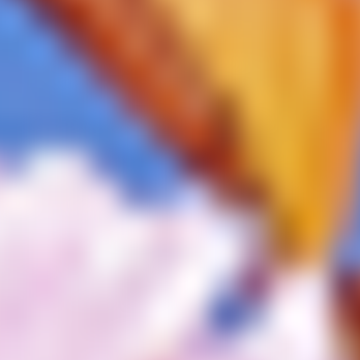} & \includegraphics[width=33mm]{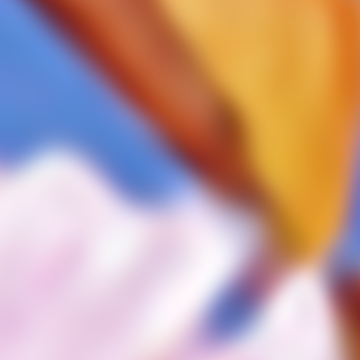} &
        \includegraphics[width=33mm]{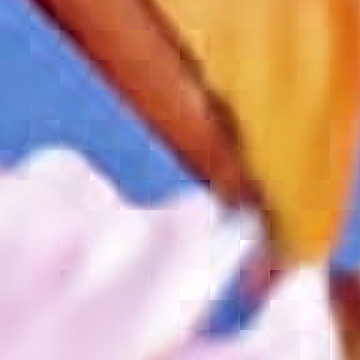} & \includegraphics[width=33mm]{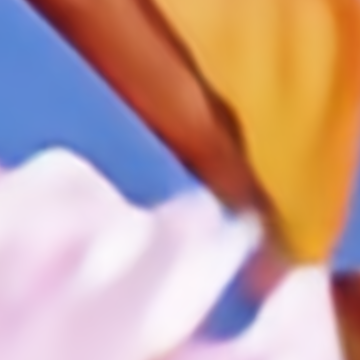} \\
        
        \includegraphics[width=4.4mm]{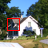}
        \includegraphics[width=17.6mm]{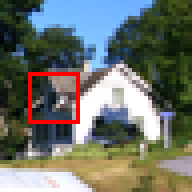} &
        \includegraphics[width=33mm]{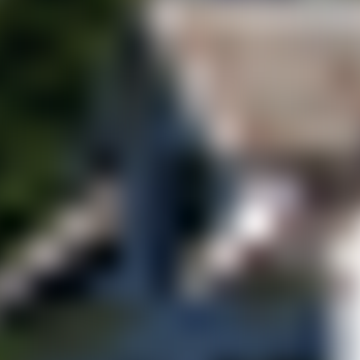} & \includegraphics[width=33mm]{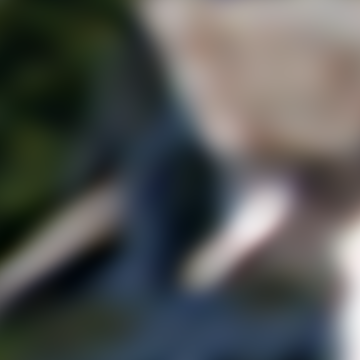} &
        \includegraphics[width=33mm]{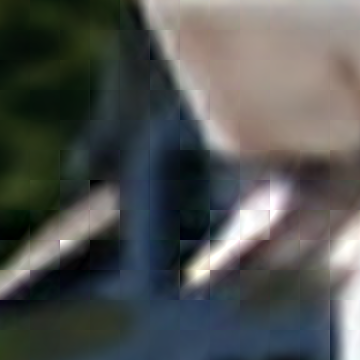} & \includegraphics[width=33mm]{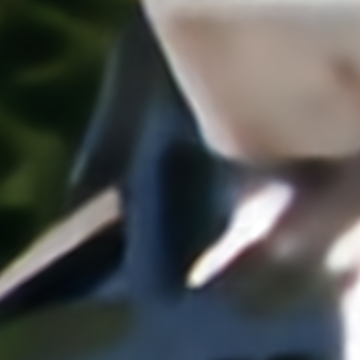} \\
        
        \begin{tabular}{c} Input (48px) \\ crop=12px \end{tabular} & Bicubic & 1-SIREN~\cite{sitzmann2020implicit} & MetaSR~\cite{hu2019meta} & LIIF (ours) 
    \end{tabular}
    \caption{\textbf{Qualitative comparison of learning continuous representation.} The input is a $48\times 48$ patch from images in DIV2K validation set, a red box indicates the crop area for demonstration ($\times30$). 1-SIREN refers to fitting an independent implicit function for the input image. MetaSR and LIIF are trained for continuous random scales in $\times1$--$\times4$ and tested for $\times30$ for evaluating the generalization to arbitrary high precision of the continuous representation.}
    %\vspace{-0.5em}
    \label{fig:div2k-x30}
\end{figure*}

\begin{figure}[]
    \centering
    \begin{tabular}{cc}
        \includegraphics[width=37mm]{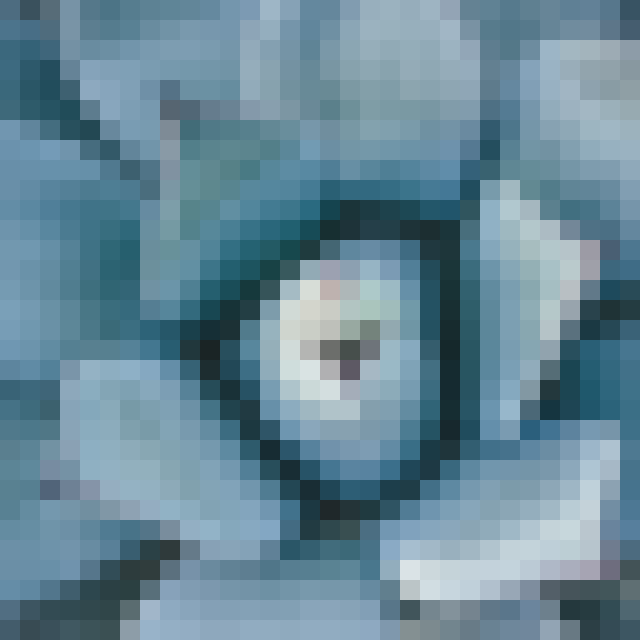} & \includegraphics[width=37mm]{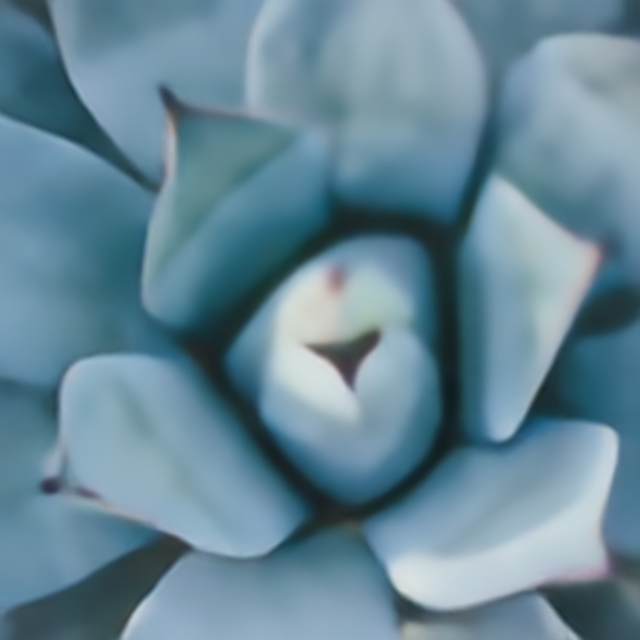} \\
        \includegraphics[width=37mm]{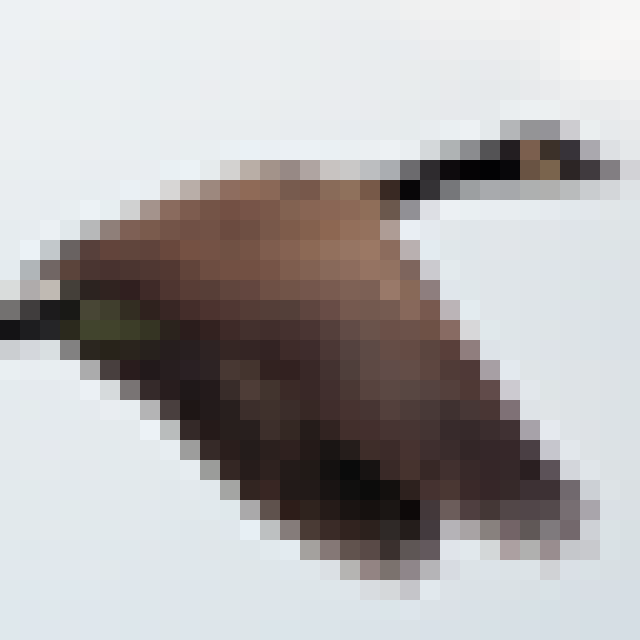} & \includegraphics[width=37mm]{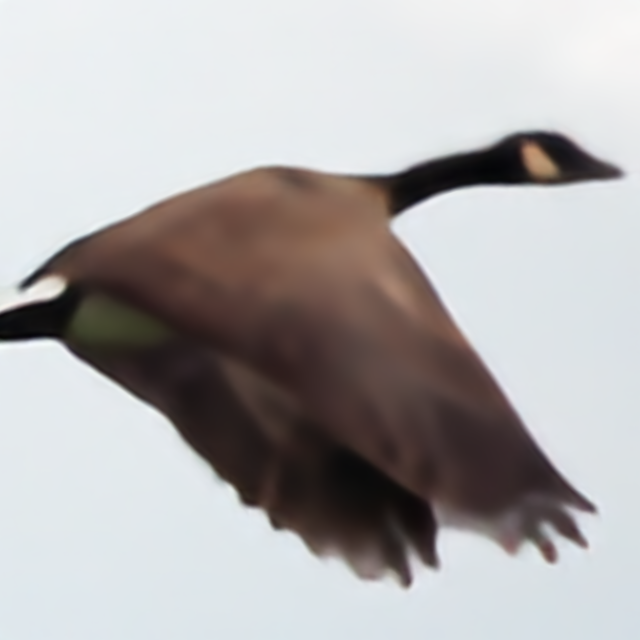} \\
        Input (32px) & LIIF (presented in 640px)
    \end{tabular}
    \caption{Demonstration of the generated LIIF representation, presented in $\times 20$ higher resolution than the input.}
    %\vspace{-0.5em}
    \label{fig:mosaic-demo}
\end{figure}

\section{Learning Continuous Image Representation}

In this section, we introduce the method for learning to generate a continuous representation for an image, an overview is demonstrated in Figure~\ref{fig:learn-liif}. Formally, in this task we have a set of images as the training set, the goal is to generate a continuous representation for an unseen image.

The general idea is to train an encoder $E_{\varphi}$ (with $\varphi$ as its parameters) that maps an image to a 2D feature map as its LIIF representation, the function $f_{\theta}$ shared by all the images is jointly trained. We hope that the generated LIIF representation is not only able to reconstruct its input, but more importantly, as a continuous representation, it should maintain high fidelity even when being presented in higher resolution. Therefore, we propose to train the framework in a self-supervised task with super-resolution.

We first take a single training image as an example, as shown in Figure \ref{fig:learn-liif}, for a training image, an input is generated by down-sampling the training image with a random scale. A ground-truth is obtained by representing the training image as pixel samples $x_{\textrm{hr}},s_{\textrm{hr}}$, where $x_{\textrm{hr}}$ are the center coordinates of pixels in the image domain, $s_{\textrm{hr}}$ are the corresponding RGB values of the pixels. The encoder $E_{\varphi}$ maps the input image to a 2D feature map as its LIIF representation. The coordinates $x_{\textrm{hr}}$ are then used to query on the LIIF representation, where $f_{\theta}$ predicts the signal (RGB value) for each of these coordinates based on LIIF representation. Let $s_{pred}$ denote the predicted signal, a training loss (L1 loss in our experiment) is then computed between $s_{pred}$ and the ground-truth $s_{\textrm{hr}}$. For batch training, we sample batches from the training set where the loss is the average over instances. We replace $x$ with $[x, c]$ when having cell decoding.

\section{Experiments}

\subsection{Learning continuous image representation}

\paragraph{Setup.}
We use DIV2K dataset~\cite{agustsson2017ntire} of NTIRE 2017 Challenge~\cite{timofte2017ntire} for experiments on learning continuous image representation. It consists of 1000 images in 2K resolution and provides low-resolution counterparts with down-sampling scales $\times2$,$\times3$,$\times4$, which are generated by imresize function in Matlab with the default setting of bicubic interpolation. We follow the standard split using 800 images in DIV2K for training. For testing, we report the results on the DIV2K validation set with 100 images which follows prior work~\cite{lim2017enhanced}, and on four standard benchmark datasets: Set5~\cite{bevilacqua2012low}, Set14~\cite{zeyde2010single}, B100~\cite{martin2001database}, and Urban100~\cite{huang2015single}.

The goal is to generate a continuous representation for a pixel-based image. A continuous representation is expected to have infinite precision that can be presented in arbitrary high resolution while maintaining high fidelity. Therefore, to quantitatively evaluate the effectiveness of the learned continuous representation, besides evaluating the up-sampling tasks of scales that are in training distribution, we propose to also evaluate extremely large up-sampling scales that are \textit{out of training distribution}. Specifically, in training time, the up-sampling scales are uniformly sampled in $\times1$--$\times4$ (continuous range). During test time, the models are evaluated on unseen images with much higher up-sampling scales, namely $\times6$--$\times30$, that are unseen scales during training. The out-of-distribution tasks evaluate whether the continuous representation can generalize to arbitrary precision.

\begin{table*}[]
    \centering
    \begin{tabular}{c|ccc|ccccc}
        \multirow{2}{*}{Method} & \multicolumn{3}{c|}{In-distribution} & \multicolumn{5}{c}{Out-of-distribution} \\
        & $\times$2 & $\times$3 & $\times$4 & $\times$6 & $\times$12 & $\times$18 & $\times$24 & $\times$30 \\
        \hline
        Bicubic~\cite{lim2017enhanced} & 31.01 & 28.22 & 26.66 & 24.82 & 22.27 & 21.00 & 20.19 & 19.59 \\
        EDSR-baseline~\cite{lim2017enhanced} & 34.55 & 30.90 & 28.94 & - & - & - & - & - \\
        EDSR-baseline-MetaSR$^\sharp$~\cite{hu2019meta} & 34.64 & 30.93 & 28.92 & 26.61 & 23.55 & 22.03 & 21.06 & 20.37 \\
        EDSR-baseline-LIIF (ours) & 34.67 & 30.96 & \textbf{29.00} & \textbf{26.75} & \textbf{23.71} & \textbf{22.17} & \textbf{21.18} & \textbf{20.48} \\
        \hline
        RDN-MetaSR$^\sharp$~\cite{hu2019meta} & 35.00 & 31.27 & 29.25 & 26.88 & 23.73 & 22.18 & 21.17 & 20.47 \\
        RDN-LIIF (ours) & 34.99 & 31.26 & 29.27 & \textbf{26.99} & \textbf{23.89} & \textbf{22.34} & \textbf{21.31} & \textbf{20.59} \\
    \end{tabular}
    \caption{\textbf{Quantitative comparison on DIV2K validation set (PSNR (dB)).} $\sharp$ indicates ours implementation. The results that surpass others by 0.05 are bolded. EDSR-baseline trains different models for different scales. MetaSR and LIIF use one model for all scales, and are trained with continuous random scales uniformly sampled in $\times1$--$\times4$.}
    \label{tab:div2k}
    %\vspace{-0.5em}
\end{table*}

\begin{table*}[]
    \centering
    \begin{tabular}{c|c|ccc|cc}
        \multirow{2}{*}{Dataset} & \multirow{2}{*}{Method} & \multicolumn{3}{c|}{In-distribution} & \multicolumn{2}{c}{Out-of-distribution} \\
        & & $\times$2 & $\times$3 & $\times$4 & $\times$6 & $\times$8 \\
        \hline
        \multirow{3}{*}{Set5} & RDN~\cite{zhang2018residual} & 38.24 & 34.71 & 32.47 & - & - \\
        & RDN-MetaSR$^\sharp$~\cite{hu2019meta} & 38.22 & 34.63 & 32.38 & 29.04 & 26.96 \\
        & RDN-LIIF (ours) & 38.17 & 34.68 & 32.50 & \textbf{29.15} & \textbf{27.14} \\
        \hline
        \multirow{3}{*}{Set14} & RDN~\cite{zhang2018residual} & 34.01 & 30.57 & 28.81 & - & - \\
        & RDN-MetaSR$^\sharp$~\cite{hu2019meta} & 33.98 & 30.54 & 28.78 & 26.51 & 24.97 \\
        & RDN-LIIF (ours) & 33.97 & 30.53 & 28.80 & \textbf{26.64} & \textbf{25.15} \\
        \hline
        \multirow{3}{*}{B100} & RDN~\cite{zhang2018residual} & 32.34 & 29.26 & 27.72 & - & - \\
        & RDN-MetaSR$^\sharp$~\cite{hu2019meta} & 32.33 & 29.26 & 27.71 & 25.90 & 24.83 \\
        & RDN-LIIF (ours) & 32.32 & 29.26 & 27.74 & \textbf{25.98} & \textbf{24.91} \\
        \hline
        \multirow{3}{*}{Urban100} & RDN~\cite{zhang2018residual} & 32.89 & 28.80 & 26.61 & - & - \\
        & RDN-MetaSR$^\sharp$~\cite{hu2019meta} & 32.92 & 28.82 & 26.55 & 23.99 & 22.59 \\
        & RDN-LIIF (ours) & 32.87 & 28.82 & \textbf{26.68} & \textbf{24.20} & \textbf{22.79} \\
    \end{tabular}
    \caption{\textbf{Quantitative comparison on benchmark datasets (PSNR (dB)).} $\sharp$ indicates ours implementation. The results that surpass others by 0.05 are bolded. RDN trains different models for different scales. MetaSR and LIIF use one model for all scales, and are trained with continuous random scales uniformly sampled in $\times1$--$\times4$.}
    \label{tab:benchmark}
    %\vspace{-0.5em}
\end{table*}

\begin{figure*}[]
    \centering
    \begin{tabular}{ccccc}
        \hspace{22mm} & \hspace{33mm} & \hspace{33mm} & \hspace{33mm} & \hspace{33mm} \\
        \multicolumn{5}{c}{\includegraphics[width=.99\linewidth]{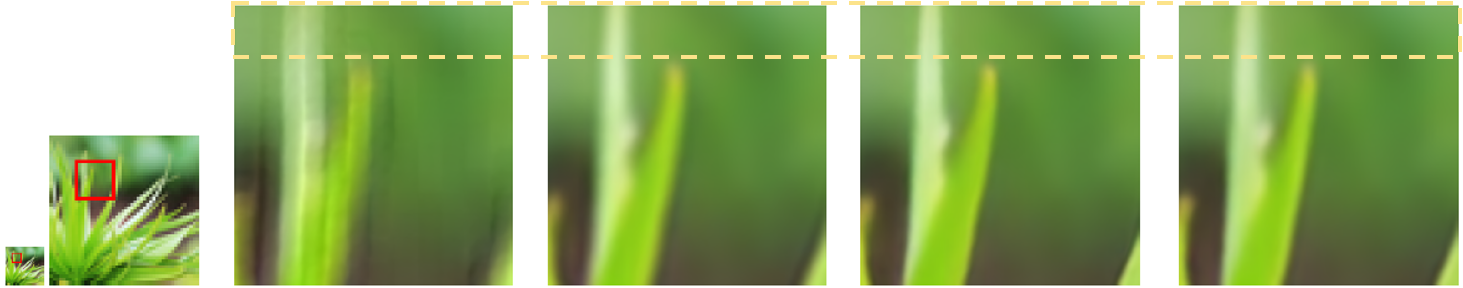}} \\
        ~~Input (48px) & ~~Cell-1/1 & ~~Cell-1/2 & ~~Cell-1/30 & No-cell
    \end{tabular}
    \caption{\textbf{Qualitative ablation study on cell decoding.} The model is trained for $\times1$--$\times4$ and tested for $\times30$. The annotation 1/$k$ refers to the cell size is 1/$k$ to a pixel in the input image. The pictures demonstrate that the learned cell generalizes to unseen scales, using a proper cell size (1/30 in this case) is less blurry (e.g. the area inside the dashed line).}
    \label{fig:div2k-abl-cell}
\end{figure*}

\begin{table*}[]
    \centering
    \begin{tabular}{c|ccc|ccccc}
        & \multicolumn{3}{c|}{In-distribution} & \multicolumn{5}{c}{Out-of-distribution} \\
        & $\times$2 & $\times$3 & $\times$4 & $\times$6 & $\times$12 & $\times$18 & $\times$24 & $\times$30 \\
        \hline
        LIIF & 34.67 & 30.96 & 29.00 & 26.75 & 23.71 & 22.17 & 21.18 & 20.48 \\
        LIIF (-c) & 34.53 & 30.92 & 28.97 & 26.73 & 23.72 & 22.19 & 21.19 & 20.51 \\
        LIIF (-u) & 34.64 & 30.94 & 28.98 & 26.73 & 23.69 & 22.16 & 21.17 & 20.47 \\
        LIIF (-e) & 34.63 & 30.95 & 28.97 & 26.72 & 23.66 & 22.13 & 21.14 & 20.45 \\
        LIIF (-d) & 34.65 & 30.94 & 28.98 & 26.71 & 23.64 & 22.10 & 21.12 & 20.42 \\
    \end{tabular}
    \caption{\textbf{Quantitative ablation study on design choices of LIIF.} Evaluated on the DIV2K validation set (PSNR (dB)). -c/u/e refers to removing cell decoding, feature unfolding, and local ensemble correspondingly. -d refers to reducing the depth of the decoding function.}
    \label{tab:div2k-abl-all}
    %\vspace{-0.5em}
\end{table*}

\vspace{-1em}
\paragraph{Implementation details.} We follow prior work~\cite{lim2017enhanced} and use $48\times 48$ patches as the inputs for the encoder. Let $B$ denote the batch size, we first sample $B$ random scales $r_{1\sim B}$ in uniform distribution $\mathcal{U}(1, 4)$, then we crop $B$ patches with sizes $\{48 r_i \times 48 r_i\}_{i=1}^B$ from training images. $48\times 48$ inputs are their down-sampled counterpart. For the ground-truths, we convert these images to pixel samples (coordinate-RGB pairs) and we sample $48^2$ pixel samples for each of them so that the shapes of ground-truths are the same in a batch.

Our method can be combined with different encoders. We use EDSR-baseline~\cite{lim2017enhanced} or RDN~\cite{zhang2018residual} (without their up-sampling modules) as the encoder $E_{\varphi}$, they generate a feature map with the same size as the input image. The decoding function $f_{\theta}$ is a 5-layer MLP with ReLU activation and hidden dimensions of 256. We follow \cite{lim2017enhanced} and use L1 loss. For training, we use bicubic resizing in PyTorch~\cite{NEURIPS2019_9015} to perform continuous down-sampling. For evaluation of scales $\times2$,$\times3$,$\times4$, we use the low resolution inputs provided in DIV2K and benchmark datasets (with border-shaving that follows \cite{lim2017enhanced}). For evaluation of scales $\times 6$--$\times 30$ we first crop the ground-truths to make their shapes divisible by the scale, then we generate low-resolution inputs by bicubic down-sampling. We use Adam~\cite{kingma2014adam} optimizer with an initial learning rate $1\cdot 10^{-4}$, the models are trained for 1000 epochs with batch size 16, the learning rate decays by factor 0.5 every 200 epochs. The experimental setting of MetaSR is the same as LIIF, except for replacing LIIF representation with their meta decoder.

\vspace{-1em}
\paragraph{Quantitative results.} In Table~\ref{tab:div2k} and Table~\ref{tab:benchmark}, we show a quantitative comparison between our method and: (i) EDSR-baseline, RDN: encoders with up-sampling modules, (ii) MetaSR~\cite{hu2019meta}: encoders with their meta decoder. EDSR-baseline and RDN reply on up-sampling modules, they are trained with different models for different scales and cannot be tested for out-of-distribution scales. For in-distribution scales, we observe that our method achieves competitive performance to prior works. Note that both EDSR-baseline and RDN are trained and evaluated for a specific scale, thus they may have more advantages on a specific task than our method. For out-of-distribution scales, both EDSR-baseline and RDN cannot be directly applied, we observe LIIF outperforms MetaSR, which shows the advantage of using implicit neural representation becomes more obvious when the scale is larger.

\vspace{-1em}
\paragraph{Qualitative results.} We demonstrate a qualitative comparison in Figure~\ref{fig:div2k-x30}. In the figure, 1-SIREN refers to independently fitting a SIREN~\cite{sitzmann2020implicit} neural implicit function for the test image, i.e., one neural network for one image without using an image encoder. MetaSR and LIIF are trained with scales $\times1$--$\times4$ and are tested for scale $\times30$. From the visualization, we observe that LIIF is significantly better than other methods. While 1-SIREN is capable of fitting an image as a neural implicit function, it does not share the knowledge across images, therefore its performance in higher precision is limited. MetaSR shows discontinuity, while LIIF is capable of demonstrating visually pleasing results, it maintains high fidelity even in an extreme $\times30$ scale that is out of training distribution.

\begin{table*}[]
    \centering
    \begin{tabular}{c|ccc|ccccc}
        & \multicolumn{3}{c|}{In-distribution} & \multicolumn{5}{c}{Out-of-distribution} \\
        & $\times$2 & $\times$3 & $\times$4 & $\times$6 & $\times$12 & $\times$18 & $\times$24 & $\times$30 \\
        \hline
        LIIF & 34.67 & 30.96 & 29.00 & 26.75 & 23.71 & 22.17 & 21.18 & 20.48 \\
        LIIF ($\times$2-only) & 34.71 & 30.72 & 28.77 & 26.51 & 23.54 & 22.05 & 21.09 & 20.40 \\
        LIIF ($\times$3-only) & 34.32 & 30.98 & 28.99 & 26.71 & 23.68 & 22.15 & 21.17 & 20.48 \\
        LIIF ($\times$4-only) & 34.08 & 30.87 & 29.00 & 26.77 & 23.74 & 22.21 & 21.21 & 20.51
    \end{tabular}
    \caption{\textbf{Quantitative ablation study on learning to generate LIIF with super-resolution for a specific up-sampling scale.} Evaluated on the DIV2K validation set (PSNR (dB)). $\times k$-only refers to training the model with the sample pairs of up-sampling scale $k$.}
    \label{tab:div2k-abl-single-scale}
    %\vspace{-0.5em}
\end{table*}

\subsection{Ablation study}

\paragraph{Cell decoding.} In cell decoding, we attach the shape of query pixel to the input of decoding function $f_\theta$, which allows the implicit function to predict different values for differently sized pixels at the same location. Intuitively, it tells $f_\theta$ to gather local information for predicting the RGB value for a given query pixel.

By comparing LIIF and LIIF (-c) in Table~\ref{tab:div2k-abl-all}, we first observe that using cell decoding improves the performance for scales $\times2$,$\times3$,$\times4$, which is expected according to our motivation. However, as the scale goes up, when LIIF is presented in out-of-distribution high resolution, it seems that cell decoding can hurt the performance of the PSNR value. Is this indicating that the learned cell does not generalize to out-of-distribution scales?

To have a closer look, we perform a qualitative study in Figure~\ref{fig:div2k-abl-cell}. The pictures are generated in a task of $\times30$ up-sampling scale, where the model is trained for $\times1$--$\times4$. Cell-1/$k$ refers to using the cell size that is 1/$k$ to a pixel in the input image. Therefore, cell-1/30 is the one that should be used for $\times$30 representation. From the figure, we can see that cell-1/30 displays much clearer edges than cell 1/1, cell-1/2, and no-cell. This is expected if we make an approximation that we assume the cell query is simply taking the average value of the image function $I^{(i)}(x)$ in the query pixel, decoding by a cell that is larger than the actual pixel size is similar to having a ``large averaging filter'' on the image. Similar reasons also apply to no-cell, since it is trained with scales $\times1$--$\times4$, it may implicitly learn a cell size for $\times1$--$\times4$ during training, which may make it blurry in $\times 30$.

In summary, we observe that using cell decoding can improve the visual results for both in-distribution and out-of-distribution large scales, but it may not be reflected on the PSNR metric for some out-of-distribution large scales. We hypothesize this is because the PSNR is evaluated towards the high-resolution ``ground-truth'', which is not a unique counterpart and the uncertainty can get much higher on larger scales.

\vspace{-1em}
\paragraph{Training with $\times k$-only.} We evaluate training LIIF with super-resolution for a fixed up-sampling scale, the results are shown in Table~\ref{tab:div2k-abl-single-scale}. As shown in the table, we observe that while training with a specific scale can achieve slightly better performance for that specific scale, it performs worse than training with random scales when evaluated on other in-distribution scales. For out-of-distribution scales, it is interesting to see that LIIF ($\times 4$-only) achieves slightly better PSNR than LIIF trained with random scales, but we hypothesize there are two potential reasons: (i) training with $\times 4$ is more biased to high scales than training with $\times 1$--$\times 4$; (ii) In LIIF ($\times 4$-only), the cell decoding is not learned (as the cell size is fixed during training), therefore, similar to the comparison between LIIF and LIIF (-c), it is possible that cell decoding improves the visual results but is not reflected on the PSNR metric for some large scales. We observe visual comparisons that are similar to Figure~\ref{fig:div2k-abl-cell} which supports (ii).

\vspace{-1em}
\paragraph{Other design choices.} In Table~\ref{tab:div2k-abl-all} we have an ablation study on other design choices. We find feature unfolding mainly helps representation in moderate scales in the comparison between LIIF and LIIF (-u). By comparing LIIF with LIIF (-e), we observe that the local ensemble consistently improves the quality of the continuous representation, which demonstrates its effectiveness. To confirm the benefits of using a deep decoding function, we compare LIIF to LIIF (-d), which reduces 5 layers to 3 layers. It turns out that having a deep decoding function is beneficial for in-distribution scales and also generalizes better to out-of-distribution scales.

\begin{table}[]
    \centering
    \begin{tabular}{c|c|cc}
        Task & Method & PSNR (dB) \\
        \hline
        $L=64,$ & Up-sampling modules~\cite{lim2017enhanced} & 34.78 \\
        $H=128$ & LIIF (ours) & \textbf{35.96} \\
        \hline
        $L=32,$ & Up-sampling modules~\cite{lim2017enhanced} & 27.56 \\
        $H=256$ & LIIF (ours) & \textbf{27.74} \\
    \end{tabular}
    \caption{\textbf{Comparison of learning with size-varied ground-truths.} Evaluated on the CelebAHQ dataset. The task is to map a face image in $L\times L$ resolution to $H\times H$ resolution, where the training images are in resolutions uniformly distributed from $L\times L$ to $H\times H$.}
    \label{tab:facesr}
    %\vspace{-0.5em}
\end{table}

\subsection{Learning with size-varied ground-truths}

In this section, we introduce further applications of LIIF continuous representation. Since LIIF representation is resolution-free, it can be compared with \emph{arbitrarily sized} ground-truths. Specifically, in the tasks where the ground-truths are images in different resolutions (which is common in practice), suppose we have a model that generates a fixed-size feature map, we do not need to resize the size-varied ground-truth to the same size \emph{which sacrifices data fidelity}. LIIF can naturally build the bridge between the fixed-size feature map and the ground-truths in different resolutions. Below we show an image-to-image task as an example of this application.

\vspace{-1em}
\paragraph{Setup.} We use CelebAHQ~\cite{karras2017progressive} dataset, that has 30,000 high-resolution face images selected from the CelebA~\cite{liu2015deep} dataset. We split 25,000 images as the training set, 5,000 images as the test set (where we use 100 images for model selection). The task is to learn a mapping from a face image in $L\times L$ (low) resolution to its counterpart in $H\times H$ (high) resolution. However, the sizes of images in the training set are uniformed distributed from $L\times L$ to $H\times H$, and for every training image, we have its down-sampled counterpart in $L\times L$ resolution (as input).

We highlight that while this problem is also a super-resolution task, it is essentially different from the super-resolution task in the previous section of learning continuous representation for pixel-based images. In previous experiments, as we assume the dataset contains general natural scenes (not category-specific), the training can be patched-based. For a specific up-sampling scale, the input and output size can be fixed for a super-resolution model since we can crop patches of any size in an image. However, in this task, we want the model to take the whole face image in $L\times L$ resolution as input to follow the input distribution during test time, instead of training in a patch-based style. Therefore, we will need to address the challenge of \emph{size-varied ground-truths}. Note that the input can potentially be any other fixed-size information (e.g. with natural noise and perturbation) for predicting the output image, we choose the input information as a $L\times L$ low-resolution counterpart here for simplicity. In general, this task is framed as an image-to-image task with size-varied ground-truths.

\vspace{-1em}
\paragraph{Methods.} We compare two end-to-end learning methods for this task. The first is denoted by ``Up-sampling modules''. In this method, since the ground-truths need to have the same resolution, all the ground-truths are resized to $H\times H$ with bicubic resizing, then an encoder is trained with up-sampling modules on the top. The second is to use LIIF representation, where we train the same encoder that generates a feature map, but we take the feature map as LIIF representation and we jointly train the encoder with the decoding function. In this case, since LIIF representation can be presented in arbitrary resolution, all the ground-truths can keep their original resolution for supervision.

\vspace{-1em}
\paragraph{Implementation details.} The $E_\varphi$ is a EDSR-baseline encoder and $f_\theta$ is a 5-layer MLP with ReLU activation and hidden dimensions of 256 (the same as previous experiments). We follow \cite{lim2017enhanced} for up-sampling modules and the training loss is L1 loss. We use Adam~\cite{kingma2014adam} optimizer, with initial learning rate $1\cdot 10^{-4}$, the models are trained for 200 epochs with batch size 16, the learning rate decays by factor 0.1 at epoch 100.

\vspace{-1em}
\paragraph{Results.} The evaluation results are shown in Table~\ref{tab:facesr}. For both tasks of $L=64,H=128$ and $L=32,H=256$, we consistently observe that using LIIF representation is significantly better than resizing the ground-truths to the same size and training with classical up-sampling modules. While the resizing operation sacrifices data fidelity, training with LIIF representation can naturally exploit the information provided in ground-truths in different resolutions. The results demonstrate that LIIF provides an effective framework for learning tasks with size-varied ground-truths.

\section{Conclusion}

In this paper, we presented the Local Implicit Image Function (LIIF) for continuous image representation. In LIIF representation, each image is represented as a 2D feature map, a decoding function is shared by all the images, which outputs the RGB value based on the input coordinate and neighboring feature vectors.

By training an encoder with LIIF representation in a self-supervised task with super-resolution, it can generate continuous LIIF representation for pixel-based images. The continuous representation can be presented in extreme high-resolution, we showed that it can generalize to much higher precision than the training scales while maintaining high fidelity. We further demonstrated that LIIF representation builds a bridge between discrete and continuous representation in 2D, it provides a framework that can naturally and effectively exploit the information from image ground-truths in different resolutions. Better architectures for the decoding function and more applications on other image-to-image tasks may be explored in future work.

\vspace{1em}
{\footnotesize \textbf{Acknowledgements.}~This work was supported, in part, by grants from DARPA LwLL, NSF 1730158 CI-New: Cognitive Hardware and Software Ecosystem Community Infrastructure (CHASE-CI), NSF ACI-1541349 CC*DNI Pacific Research Platform, and gifts from Qualcomm and TuSimple.}

{\small
\bibliographystyle{ieee_fullname}
\bibliography{main}
}

\end{document}